# Multi-time-horizon Solar Forecasting Using Recurrent Neural Network

Sakshi Mishra
Transmission Planning Engineer, West Transmission Planning
American Electric Power
Tulsa, USA
sakshi.m@outlook.com

Praveen Palanisamy
Alumni, Robotics Institute
Carnegie Mellon University
Pittsburgh, USA
praveen.palanisamy@outlook.com

*Abstract*— The non-stationarity characteristic of the solar power renders traditional point forecasting methods to be less useful due to large prediction errors. This results in increased uncertainties in the grid operation, thereby negatively affecting the reliability and increased cost of operation. This research paper proposes a unified architecture for multi-time-horizon predictions for short and long-term solar forecasting using Recurrent Neural Networks (RNN). The paper describes an end-to-end pipeline to implement the architecture along with methods to test and validate the performance of the prediction model. The results demonstrate that the proposed method based on the unified architecture is effective for multi-horizon solar forecasting and achieves a lower root-mean-squared prediction error compared to the previous best performing methods which use one model for each time-horizon. The proposed method enables multi-horizon forecasts with real-time inputs, which have a high potential for practical applications in the evolving smart grid.

*Keywords*— *Artificial Neural Network, Forecasting, Predictive Analysis, Recurrent Neural Network, Renewable Energy, Solar Power, Multi-time-horizon Solar Forecasting, Smart Grid*

## I. INTRODUCTION

Today's power grid has become dynamic in nature mainly because of three changes in the modern grid: 1. Higher penetration level of renewables, 2. Introduction and rapidly increasing deployment of storage devices, and 3. Loads becoming active by participating in demand response. This dynamic modern grid faces the challenge of strong fluctuations due to uncertainty. There is a critical need of gaining real time observability, control, and improving renewable generation forecast accuracy to enhance the resiliency and to keep the operational costs sustainable. Independent system operators (ISOs) have already been facing challenges with higher renewable penetration on the grid due to the uncertainties resulting from short-term forecasting errors. In the year 2016, California ISO doubled its frequency regulation service requirements, causing a sharp rise in the cost of requirements, to manage the recurring short-term forecasting errors in renewable generation [1]. The Western Electricity Coordinating Council (WECC) could save $5 billion per year by integrating wind and solar forecasts into unit commitment, according to the study conducted by Lew et al [2]. Thus, it is clear that the increased grid penetration levels of solar with its inherent variability caused by a combination of intermittence, high-frequency and non-stationarity, poses problems for grid reliability and increases the grid operation costs at various time-scales. For example, day-ahead solar forecast accuracy plays a significant role in the effectiveness of Unit Commitment (UC); very-short-term solar forecast errors due to fluctuations caused by the passing clouds lead to sudden changes in PV plant outputs that can cause strain to the grid by inducing voltage-flickers and real-time balancing issues. Thus, solar power generation forecasting is an area of paramount research, as the need for robust forecast for all timescales (weekly, day-ahead, hourly and intra-hour) is critical for effectively incorporating the increasing amount of solar energy resources at a global level and contributing to the evolution of the smart grid. Moreover, improving the accuracy of solar forecasting is one of the low cost methods for efficiently integrating solar energy into the grid.

The rest of the paper is organized as follows. The literature is reviewed and the significant shortcomings of the current forecasting approaches are recognized in Section II. Section II further introduces the capabilities of the proposed unified architecture and the novel algorithm to fill in the gap between the need to improve the forecasting techniques and the existing approaches. Section III introduces the proposed unified architecture based on RNNs and the training procedure used for implementing the neural network. Exploratory data analysis, evaluation metric and the structure of the input data, and the proposed algorithm are presented in Section IV. Section V discusses the results and their interpretation. The paper is concluded with Section VI, which also identifies the future avenue of research in this method of solar forecasting.

## II. LITERATURE REVIEW AND PROBLEM IDENTIFICATION

Forecasting methods which have been used for renewable generation and electric load forecasting prediction can be mainly classified into five categories: 1) Regressive methods, such as Autoregressive (AR), AR integrated moving average (ARIMA), and exponential smoothing (ES) models [3], [4], [5], nonlinear stationary models; 2) Artificial Intelligence (AI) techniques, such as Artificial Neural Networks (ANN) [6]-[10], k-nearest neighbors [11]-[14], fuzzy logic systems (FLSs) [15]-[17]; 3) Numerical Weather Prediction (NWP) [18]; 4) Sensing (remote and local) [19]. 5) Hybrid models, such as neuro-fuzzy systems [20]-[21], ANN and satellite derived cloud indices [22], to name a few.

Numerical Weather Prediction (NWP) models are based on physical laws of motion and thermodynamics that govern the weather. For the places where ground data is not available,

NWP models are powerful tools to forecast solar radiation. However, they pose significant limitations in predicting the precise position and extent of cloud fields due to their relatively coarse spatial resolution. Their inability to resolve the micro-scale physics associated with cloud formation renders them with relatively large error in terms of cloud prediction accuracy. In order to mitigate this limitation, NWPs are simulated at regional level (called Regional NWP) models, downscaled to derive improved site-specific forecasts. NWP has another limitation of temporal-resolution. The timescale of output variables of NWP models is from 3 hour - 6 hours for the Global Forecast System (GFS) and 1-hour (for mesoscale model), which is not useful for predicting the ramp-rate and very-short-term output fluctuations.

For the areas where the previous ground-based measurement are not available, satellite based irradiance measurement proved to be a useful tool [22]. The images from satellite are used to analyze the time evolution of air mass by the superimposition of images of the same area. Radiometer installed in the satellite records the radiance, states of the atmosphere (clear sky to overcast) impacts the radiance. Satellite sensing has the main limitation of determining an accurate set point for the radiance value under clear sky conditions and under dense cloudiness condition from every pixel in every image. Another limitation of solar irradiance forecasting using remotes sensing with satellite is the algorithms that are classified as empirical or statistical [23]-[24]. These algorithms are based on simple statistical regression between surface measurements and satellite information and do not need accurate information of the parameters that model the solar radiation attenuation through the atmosphere. So the ground-based solar data is required for these satellite statistical algorithms anyway.

The aforementioned limitations of NWP and sensing models have steered the short-term solar forecasting research towards time-series analysis using statistical models and more recently AI techniques. Statistical techniques can mainly be classified as [25]: 1) Linear stationary models (Autoregressive models, Moving Average models, Mixed Autoregressive Moving Average Models, and Mixed Autoregressive Moving Average models with exogenous variables); 2) Nonlinear stationary models; 3) Linear non-stationary models (Autoregressive integrated moving average models and Autoregressive integrated moving average models with exogenous variables). Though these conventional statistical techniques provide a number of advantages over NWP and sensing methods, but these are often limited by strict assumptions of normality, linearity, variable independence. Artificial Neural Networks (ANN) are able to represent complex non-linear behaviors in higher dimensional settings. When exogenous variables like humidity, temperature and pressure are considered in the process of solar forecasting - ANNs act as universal function approximators to model the complex non-linear relationships between these variables and their relationship with the Global Horizontal Irradiance (GHI). An ANN with multiple hidden layers can be called A Deep Neural Network (DNN). With the advancements in computational capabilities, DNNs have proven to be effective and efficient in solving complex problems in many fields including image recognition, automatic speech recognition and natural language processing etc [26]. Although, feed-forward neural network models have been used for solar forecasting problem, the use of Recurrent Neural Networks (RNN) models have not been explored yet, to the best of the author's knowledge. RNN is a class of ANN that capture the dynamics of sequences using directed cyclic feedback connections [27]. Feedforward neural networks rely on the assumption of independence among the data points or samples. The entire state of the network is lost after processing each data point (sample). Unlike vanilla feedforward neural networks, recurrent neural networks (RNNs) exhibit dynamic temporal behavior by using their internal memory to process arbitrary sequences if inputs, which can be harnessed in predicting the irradiance for the next time step by considering the input from many previous times steps. Recent advances in parallelism, network architectures, optimization techniques, and graphics processing units (GPUs) have enabled successful large-scale learning with RNNs overcoming their traditional limitations of being difficult to train due to having millions of parameters.

Several methods have been proposed for solar forecasting in the past but most of them were modeled for a particular time-horizon and no single model performed well compared to others for multi-time-horizon forecasting/prediction. In addition, the state-of-the-art methods used for solar forecasting primarily focuses on averaged rather than instantaneous forecasts. This paper proposes two approaches using RNNs. 1). A single system that is capable of being trained to output solar forecast for 1-hour or 2-hour or 3-hour, or for 4-hour time horizons. ii) A unified architecture that can predict/forecast the solar irradiance for multi-time-horizons; for example, the trained model can predict/forecast the solar irradiance values for the 1-hour, 2-hour, 3-hour and 4-hour time horizons. Our proposed method is capable of taking a time-series data as the input and provides predictions with a forward inference time in the order of milliseconds, enabling real-time forecasts based on live measured data. This offers a great value for industrial applications that require real-time multi-time-horizon forecasting for overcoming the current operational challenges with high penetration of renewable source of energy.

III. ARCHITECTURE AND ALGORITHM

The RNN resembles a feedforward neural network except for additional directed edges. These edges span adjacent time steps, introducing the notion of temporal component to the model. Theoretically, the RNN architecture enables the network to make use of past information in sequential data.

*A. Recurrent Neural Network – Unified Architecture*

The input to an RNN is a sequence, and its target can be a sequence or a single value. An input sequence is denoted by $(x^{(1)}, x^{(2)},...x^{(T)})$, where each sample/data-point $x^{(t)}$ is a real valued vector. The target sequence is denoted by $(y^{(1)}, y^{(2)},...y^{(T)})$ and the predicted target data-point is denoted by

($y^{(1)}$, $y^{(2)}$,...$y^{(T)}$). There are three dimensions to the input of the RNN (shown in Figure 1): 1) Mini-batch Size; 2) Number of columns in the vector per time-step; and 3) Number of time-steps. Mini-batch size is the sample length (data-points in the time-series). Number of columns are the input features in the input vector. The number of time-steps is the differentiating factor of RNN, which unfolds the input vector over time.

In a typical multilayer feedforward neural network, the input vector is fed to the neurons at the input layer, which then gets multiplied by the activation function to produce the intermediate output of the neuron, this output then becomes the input to the neuron in the next layer. The net input (denoted by *input_sum$_i$*) to this neuron belonging to the next layer is the weight on connections (*W*) multiplied by previous neuron's output with the bias term, as shown in Equation 1. An activation function (denoted by g) is then applied to the *input_sum$_i$* to produce the output from the neuron Equation 2 and 3.

$$input\_sum_i = W_i \cdot x_i + b \quad (1)$$
$$a_i = g(input\_sum_i) \quad (2)$$
$$a_i = g(W_i \cdot x_i + b) \quad (3)$$

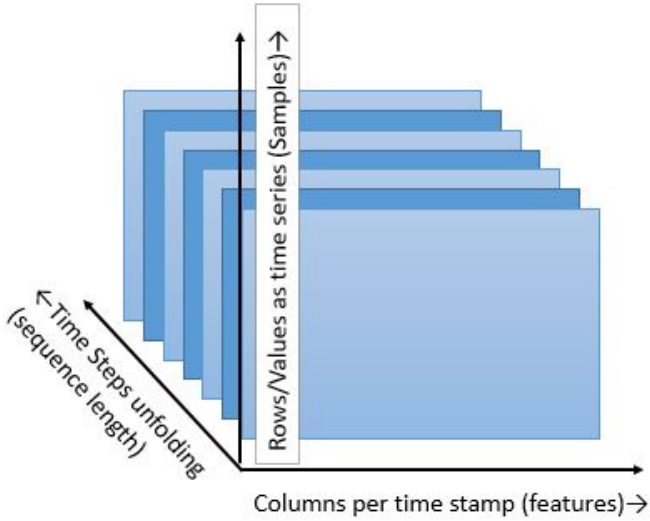

**Figure 1 Input representation for the Recurrent Neural Network**

For RNN network, at time t, neurons with recurrent edges receive input from the current sample $x^{(T)}$ and also from hidden node values $h^{(t-1)}$ in the network's previous state (Equation 4). Given the hidden node values $h^{(t)}$ at time t, the output $y^{(T)}$ at each time *t* is calculated (Equation 5).

$$h(t) = \sigma(W^{hx}x^{(t)} + W^{hh}h^{(t-1)} + b_h) \quad (4)$$

$$\hat{y}^{(t)} = g(W^{yh}h^{(t)} + b_y) \quad (5)$$

where, $W_{hx}$ is the conventional weight matrix between the input and the hidden layer and $W_{hh}$ is the recurrent weights matrix between the hidden layer and itself at adjacent time steps. $b_h$ and $b_y$ are bias parameters. The proposed architecture uses Rectified Linear Units (ReLU) as the activation function. The network unfolds the given input at time t as shown in Figure 1.

### B. Algorithm description

The unfolded network is trained across the time steps using an algorithm called backpropagation through time (BPTT) [28]. The loss function used for this regression problem is Mean Squared Loss (MSE). The loss function finds the error between the target output and the predicted output from the network. Gradients are computed using back-propagation-through time [28] and the stochastic gradient descent optimizer is used to update the weights so as to minimize the loss. The RMSE is calculated for benchmarking purposes.

The motivation to use a RNN is to identify and learn the complex relationship between sequences of various exogenous variables and their combined impact on the solar irradiance. This, in the author's view, enables the algorithm to recognize non-linear contributing factors for example the atmospheric conditions, which may lead to cloud formation in nearby time horizon. This is one of the reasons the prediction RMSE is lower in the proposed approach compared to other reported approaches.

### C. Solar Forecasting – input features and predictions

The algorithm and the unified architecture developed in this paper were trained and tested on data from the NOAA's SURFRAD [31] sites similar to the previous works in the literature [29][32]. The input features are: downwelling global solar (Watts/m^2), upwelling global solar (Watts/ m^2), direct-normal solar (Watts/ m^2), downwelling diffuse solar (Watts/ m^2), downwelling thermal infrared (Watts/ m^2), downwelling IR case temp. (K), downwelling IR dome temp. (K), upwelling thermal infrared (Watts/ m^2), upwelling IR case temp. (K), upwelling IR dome temp. (K), global UVB (milliWatts/ m^2), photosynthetically active radiation (Watts/ m^2), net solar (dw_solar - uw_solar) (Watts/ m^2), net infrared (dw_ir - uw_ir) (Watts/ m^2), net radiation (netsolar+netir) (Watts/ m^2), 10-meter air temperature (C), relative humidity (%), wind speed (ms^1), wind direction (degrees, clockwise from north), and station pressure (mb). According to Dobbs [29], Global downwelling solar measurements best represent the Global Horizontal Irradiance (GHI) at the SURFRAD sites, which was validated in this paper through exploratory data analysis. [Figure 3] shows the daily averages of Clear Sky GHI and global downwelling solar at SURFRAD site for a year, both the variable follow the same trend. [Figure 4] shows that both these variables are positively correlated.

Bird model is used to calculate clear sky GHI [30]. At time t, clear sky GHI is denoted by $GHI_{clear}^t$, representing the theoretical GHI at time *t* assuming zero cloud coverage. At time *t*, the ratio between the instantaneously observed $GHI^t$ and the theoretical maximum $GHI_{clear}^t$ is called clear sky index, denoted by $Kt_i^{(t)}$, this parameter is introduced in [29].

$Kt_i^{(t)}$ is used as the dependent variable for training and testing the model. $Kt_i^{(t)}$ is averaged over forecasting horizon, for hourly predictions. The averaged hourly clear sky index ending at time f.h. is denoted by $Kt_a^{(f.h.)}$ and calculated as shown in Equation 6.

$$\frac{\sum_{s=f.h.-60}^{f.h.} Kt_i^{(s)}}{60} \quad (6)$$

## IV. EXPLORATORY DATA ANALYSIS AND PROPOSED ALGORITHM

### A. Exploratory Data Analysis

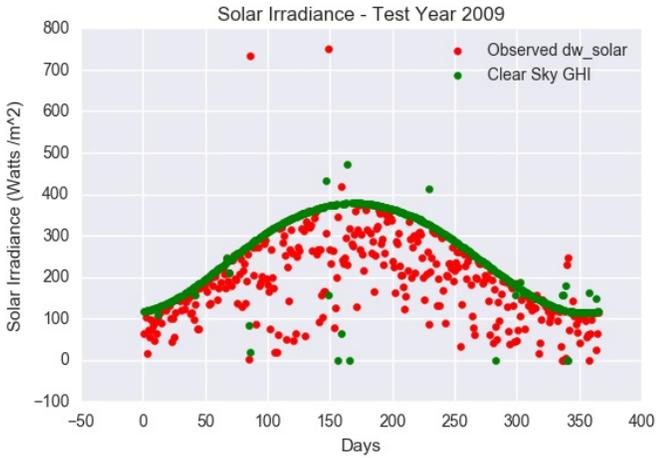

Figure 2 Cleary Sky and Observed Irradiance

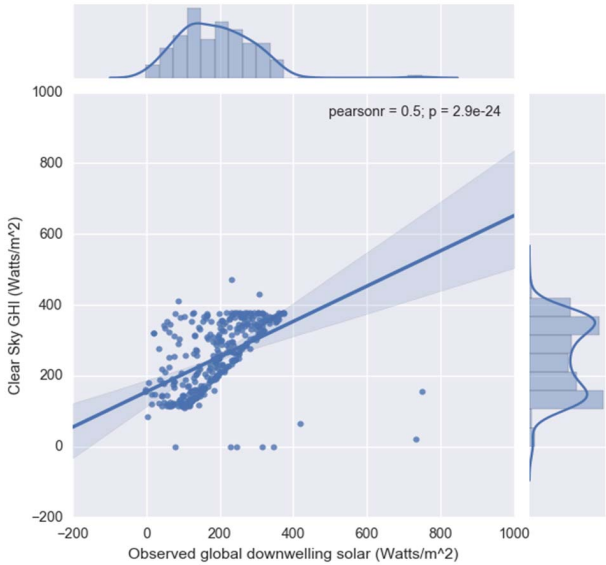

Figure 3 Correlation between Observed and Clear Sky GHI

Figure[2] shows the variation of the observed global downwelling solar and clear sky global horizontal irradiance for the year 2010 at Boulder, CO. Figure [3] shows the correlation between the two for same year and same site. There is a positive and strong correlation between the two quantities, as shown by the regression line plotted on to the scatter plot.

### B. Evaluation Metric

The algorithm uses Mean Squared Error (MSE) as a measure to find the difference between the target and the output that the neural network produces during the training process, this is shown in Equation 7. Later in the process, for the purpose of benchmarking, the Root Mean Squared Error is calculated by taking square root of the MSE values.

$$MSE = \frac{1}{n}\sum_{i=1}^{n}(Y_i - \hat{Y}_i)^2 \quad (7)$$

Where $Y$ is a vector of target values and $\hat{Y}$ is a vector of n predicted values.

### C. Algorithm

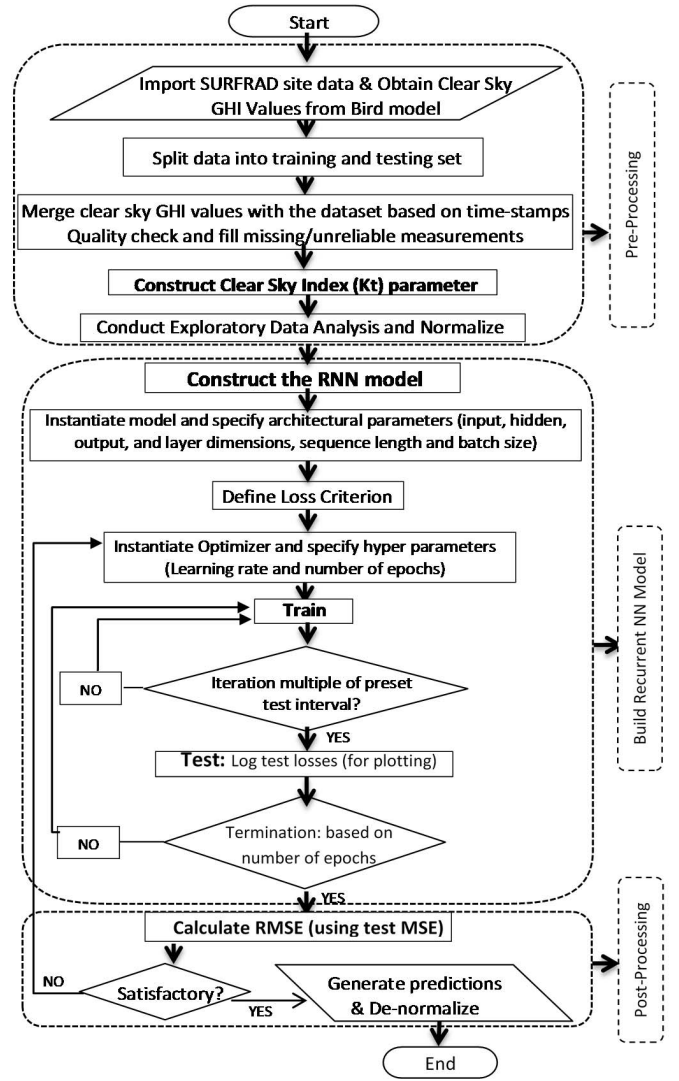

Figure 4 Flow chart of the proposed method

The flowchart of the proposed Unified Recurrent Neural Network Architecture based method is shown in Figure 4. The overall algorithm can be divided into three mail blocks:

*1) Preprocessing*

The site-specific data is imported and clear sky global horizontal irradiance values for that site are obtained from the Bird Model. The two are merged. Dataset is split into train and testing sets. The clear sky index parameter is created as the ratio of observed global downwelling solar (Watts /m^2) and GHI (Watts /m^2). Kt is a dimensionless parameter. The missing values in the data are replaced by the mean and/or the neighborhood values. Exploratory data analysis is conducted to identify and eliminate extreme outliers in order to normalize the data.

*2) RNN training and testing*

A Recurrent Neural Network based model architecture is instantiated by specifying the architectural parameters: input dimension (number of nodes at the input layer; 22), hidden dimension (number of nodes in the hidden layer; 15), layer dimension (number of hidden layers; 1) and output dimension (number of nodes in the output layer; 4). Sequence length, which unfolds as time-steps is also defined here along with the batch size. The model is trained and test by iterating through the whole dataset based on pre-set number of epochs. We used a batch size of 100 for 1000 number of epochs for obtaining the results discussed in this paper.

*3) Post-processing*

Once the training and testing is over, the stored MSE is first de-normalized and then it is used to calculate RMSE. If the RMSE is not satisfactory the hyperparameters (learning rate and number of epochs) are tuned and the model is trained again. When a satisfactory (or as expected) RMSE is achieved, the training process of the algorithm terminates.

## V. RESULTS AND DISCUSSION

The algorithm is trained using the data for the year 2010 and 2011 from the SURFRAD observations sites in Boulder, CO; Desert Rock, NV; Fort Peck, MT; Sioux Falls, SD; Bondville, IL; Goodwin Creek, MS; and Penn State, PA. The test year for each respective site was chosen to be 2009 for the purpose of benchmarking against [29] and other previously reported results in the literature. Results from the two methods proposed in this paper are presented in the following two sub-sections.

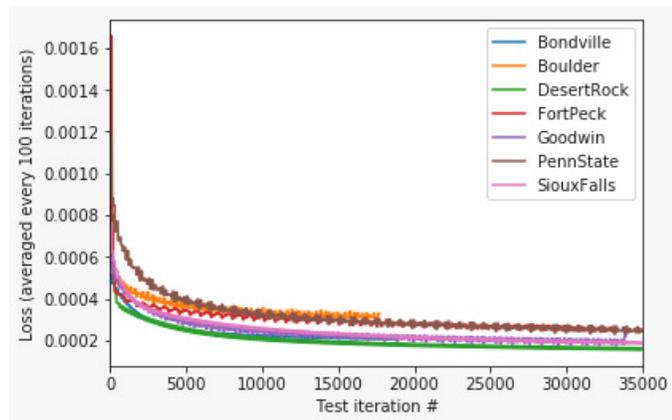

**Figure 5 Test Mean Squared Error Plot**

### A. Fixed Time Horizon Predictions

The first method uses the proposed RNN architecture and algorithm to predict for 1-hour, 2-hour, 3-hour and 4-hour time horizons, independently. In other words, four independent models are developed for 1-hour, 2-hour, 3-hour and 4-hour predictions for each of the seven SURFRAD sites. Figure 5 shows the test loss over 1000 epochs (with a batch size of 100 and test set of 3300 samples from the test year 2009) for all the seven sites. The RMSE values in Table 1 show that the proposed architecture and algorithm has lower RMSE values for all four forecasting horizons and all the seven sites, compared to the best RMSE values reported in [29] from a suite of other machine learning algorithms (Random Forests, Support Vector Machine, Gradient Boosting and vanilla Feed-Forward).

### B. Multi-time-horizon prediction

In this method, the architecture predicts for all four forecasting time horizons (1-hour, 2-hour, 3-hour, and 4-hour) in parallel; i.e. one model per SURFRAD site is developed which makes predictions for all the four time horizons. This method is the multi-time-horizon implementation of the proposed architecture. None of the methods discussed in the literature have been shown to be capable of producing multi-time-horizon predictions. Table 2 enlists the RMSE values obtained for the test years 2009, 2015, 2016 and 2017. To quantify the overall performance of the predictive model in terms of its combined forecasting accuracy for all four forecasting horizons, the mean of the RMSE values is calculated. Although, even the best RMSE values reported in the literature (for example in [29][32]) were for a single time horizon forecast at a time, the proposed method achieves a significantly lower RMSE in predictions across all the short-term (1 hour to 4 hours) forecasting time-horizons as seen in Table 2.

Capability to predict for multi-time-horizons makes the proposed method very relevant for industry applications. The real-time data can be fed to the RNN and due to its lower forward inference time, predictions can be made for multiple time horizons. The proposed method is implemented using PyTorch and the code and additional information can be found on this site: http://sakshi-mishra.github.io/.

## VI. CONCLUSION AND FUTURE WORK

Short-term solar forecasting is of great importance for optimizing the operational efficiencies of smart grids, as the uncertainties in the power systems are ever-increasing, spanning from the generation arena to the demand-side domain. A number of methods and applications have been developed for solar forecasting, with some level of predictive success. The main limitation of the approaches developed so far is their specificity with a given temporal and/or spatial resolution. For predictive analysis problems, the field of AI has become promising with the recent advances in optimization techniques, parallelism, and GPUs. AI (especially deep neural networks) thrives on data, and with decreasing cost of sensor and measurement equipment, plethora of solar data is getting available. Data availability is only going to keep increasing in the coming years. The proposed novel Unified Recurrent Neural Network Architecture harnesses the power of AI to form a high-fidelity solar forecasting engine. This architecture has the potential to be implemented as a complete forecasting system, which spans the entire spectrum of spatial and temporal horizons with a capability to take real-time data as input to produce multi-time-scale (intra-hour, hourly and day-ahead scales) predictions. In addition, the proposed algorithm outperforms traditional Machine Learning methods in terms of quality of the forecast and its low forward inference time makes it a robust real-time solar forecasting engine.

Although a deeper neural network will have more capacity, we experimentally observed that it leads to high variance in the model and therefore a reduced generalization power for the particular problem dealt in this paper. The performance of the proposed method can be further improved in several ways including hyper-parameter tuning and architectural changes like the activation functions used or the type of layers. Extension of the proposed architecture with LSTM cells and intra-hour forecasting horizons are potential future research avenues in this domain.

### Table 1
### Method 1 (Fixed Time Horizon Predictions) Results

| Year 2009 | Bondville | | Boulder | | Desert Rock | | Fort Peck | | Goodwin Creek | | Penn State | | Sioux Falls | |
|---|---|---|---|---|---|---|---|---|---|---|---|---|---|---|
| F.H. | RNN | ML | RNN | ML | RNN | ML | RNN | ML | RNN | ML | RNN | ML | RNN | ML |
| 1-hour | 16.8 | 62 | 17 | 74 | 41.7 | 52 | 21.2 | 56 | 24.8 | 71 | 8.64 | 67 | 27.2 | 52 |
| 2-hour | 20.73 | 98 | 20.7 | 108 | 57.23 | 72 | 29.7 | 81 | 25.2 | 103 | 10.5 | 97 | 32.1 | 81 |
| 3-hour | 18.78 | 116 | 21.2 | 123 | 60.54 | 83 | 25.5 | 94 | 26.9 | 125 | 11.8 | 114 | 30.6 | 96 |
| 4-hour | 17.98 | 121 | 22.9 | 125 | 49.71 | 82 | 29.4 | 93 | 22 | 120 | 10.7 | 117 | 35.3 | 103 |
| Mean RMSE | 18.57 | 99.25 | 20.45 | 107.5 | 52.29 | 72.25 | 26.45 | 81 | 24.73 | 104.8 | 10.41 | 98.75 | 31.3 | 83 |

### Table 2
### Method 2 (Multi-time-horizon prediction) Results

| Year | Forecast Horizon | Bondville | Boulder | Desert Rock | Fort Peck | Goodwin Creek | Penn State | Sioux Falls |
|---|---|---|---|---|---|---|---|---|
| 2009 | 1-hour | 0.706 | 0.593 | 0.469 | 0.709 | 0.647 | 0.679 | 0.686 |
| | 2-hour | 1.119 | 1.055 | 0.984 | 1.352 | 1.173 | 1.134 | 1.247 |
| | 3-hour | 4.524 | 4.010 | 4.527 | 6.311 | 4.586 | 3.161 | 5.061 |
| | 4-hour | 52.130 | 49.453 | 153.270 | 69.987 | 58.984 | 28.396 | 73.409 |
| | Mean RMSE | 14.619 | 13.779 | 39.812 | 19.589 | 16.347 | 8.342 | 20.100 |
| 2015 | 1-hour | 0.717 | 0.559 | 0.437 | 0.689 | 0.643 | 0.734 | 0.680 |
| | 2-hour | 1.166 | 1.026 | 0.952 | 1.302 | 1.190 | 1.180 | 1.252 |
| | 3-hour | 4.719 | 4.006 | 4.268 | 5.845 | 4.764 | 3.118 | 5.216 |
| | 4-hour | 57.839 | 49.349 | 85.952 | 57.687 | 35.814 | 25.796 | 50.823 |
| | Mean RMSE | 16.110 | 13.735 | 22.902 | 16.381 | 10.603 | 7.707 | 14.493 |
| 2016 | 1-hour | 0.731 | 0.593 | 0.456 | 0.726 | 0.669 | 0.738 | 0.723 |
| | 2-hour | 1.184 | 1.079 | 0.979 | 1.372 | 1.204 | 1.187 | 1.331 |
| | 3-hour | 4.639 | 4.227 | 4.685 | 6.406 | 4.760 | 3.174 | 5.781 |
| | 4-hour | 69.837 | 89.990 | 90.553 | 73.523 | 45.026 | 21.994 | 56.113 |
| | Mean RMSE | 19.098 | 23.973 | 24.169 | 20.507 | 12.915 | 6.773 | 15.987 |
| 2017 | 1-hour | 0.744 | 0.593 | 0.444 | 0.726 | 0.649 | 0.711 | 0.721 |
| | 2-hour | 1.190 | 1.059 | 0.949 | 1.368 | 1.179 | 1.149 | 1.323 |
| | 3-hour | 4.577 | 4.010 | 4.319 | 6.245 | 4.593 | 3.211 | 5.715 |
| | 4-hour | 81.434 | 49.453 | 44.589 | 114.072 | 32.431 | 29.307 | 59.761 |
| | Mean RMSE | 21.986 | 13.779 | 12.575 | 30.603 | 9.713 | 8.594 | 16.879 |